\documentclass[conference]{IEEEtran}
\makeatletter
\newcommand{\newlineauthors}{%
  \end{@IEEEauthorhalign}\hfill\mbox{}\par
  \mbox{}\hfill\begin{@IEEEauthorhalign}
}
\makeatother

\IEEEoverridecommandlockouts

\usepackage{cite}
\usepackage{amsmath,amssymb,amsfonts}
\usepackage{algorithm}
\usepackage{float}
\usepackage{algorithmic}
\usepackage{graphicx}
\usepackage{textcomp}
\usepackage{xcolor}
\usepackage{float}
\usepackage[bookmarks=true, hidelinks]{hyperref}
\usepackage{subcaption}
\usepackage{tabularx}
\usepackage{tabularray}
\usepackage{changepage}

\def\BibTeX{{\rm B\kern-.05em{\sc i\kern-.025em b}\kern-.08em
    T\kern-.1667em\lower.7ex\hbox{E}\kern-.125emX}}
\begin{document}

\title{Context Aware Mamba-based Reinforcement Learning for Social Robot Navigation}

\author{\IEEEauthorblockN{Syed Muhammad Mustafa}
\IEEEauthorblockA{\textit{Computer Engineering} \\
\textit{Habib University} \\
Karachi, Pakistan \\
sm06554@st.habib.edu.pk}
\and
\IEEEauthorblockN{Zain Ahmed Usmani}
\IEEEauthorblockA{\textit{Computer Science} \\
\textit{Habib University} \\ 
Karachi, Pakistan \\
zu06777@st.habib.edu.pk}
\and
\IEEEauthorblockN{Omema Rizvi}
\IEEEauthorblockA{\textit{Computer Science} \\
\textit{Habib University} \\
Karachi, Pakistan \\
or06360@st.habib.edu.pk}

\newlineauthors
\IEEEauthorblockN{Abdul Basit Memon}
\IEEEauthorblockA{\textit{Electrical \& Computer Engineering} \\
\textit{Habib University} \\ 
Karachi, Pakistan \\
basit.memon@sse.habib.edu.pk}
\and
\IEEEauthorblockN{Muhammad Mobeen Movania}
\IEEEauthorblockA{\textit{Computer Science} \\
\textit{Habib University}\\
Karachi, Pakistan\\
mobeen.movania@sse.habib.edu.pk}}

\maketitle
\begin{abstract}
Social robot navigation (SRN) is a relevant problem that involves navigating a pedestrian-rich environment in a socially acceptable manner. It is an essential part of making social robots effective in pedestrian-rich settings. The use cases of such robots could vary from companion robots to warehouse robots to autonomous wheelchairs. In recent years, deep reinforcement learning has been increasingly used in research on social robot navigation.  
Our work introduces CAMRL (Context-Aware Mamba-based Reinforcement Learning). Mamba is a new deep learning-based State Space Model (SSM) that has achieved results comparable to transformers in sequencing tasks. CAMRL uses Mamba to determine the robot's next action, which maximizes the value of the next state predicted by the neural network, enabling the robot to navigate effectively based on the rewards assigned.
We evaluate CAMRL alongside existing solutions (CADRL, LSTM-RL, SARL) using a rigorous testing dataset which involves a variety of densities and environment behaviors based on ORCA and SFM, thus, demonstrating that CAMRL achieves higher success rates, minimizes collisions, and maintains safer distances from pedestrians. This work introduces a new SRN planner, showcasing the potential for deep-state space models for robot navigation.
\end{abstract}


\section{Introduction}
Mobile robots are increasingly being deployed in pedestrian-dense environments such as shopping centers, restaurants, convention halls, and warehouses. In these crowded settings, effective navigation requires that the robot generates collision-free paths while avoiding numerous dynamic agents. an agent is any entity in the environment that has the agency to make decisions that change the state of the system. Additionally, it is desirable that the robot behavior in such environments adheres to human social norms, ensuring seamless integration. The challenge is compounded by limited communication between the robot and surrounding agents, preventing direct query of intent (an agent's intention refers to its desired goal position and preferred velocities). To address this, it is often necessary for the robot to model the behavior of each agent in the crowd, as well as its interactions with others, enabling the path planner to devise collision-free paths based solely on observable agent behavior.

The problem is complex as the communication between the robot and the other agents in the crowd, such as humans, is limited. This means that the robot's path planning module must plan a collision-free path using only the observable information available, i.e. intentions (goal position and preferred velocities) of agents would be hidden from the robot.

Reaction-based collision avoidance algorithms, such as Optimal Reciprocal Collision Avoidance (ORCA) \cite{orca} and the Social Force Model (SFM) \cite{sfm}, are widely used for this problem. These algorithms employ a one-step look-ahead strategy to avoid both static and dynamic obstacles at each time step. While computationally efficient, this approach often results in myopic and oscillatory behaviors \cite{Chen2017}. In contrast, recent advancements in deep reinforcement learning (DRL) have demonstrated its capacity to solve complex tasks, sometimes surpassing human performance. DRL holds significant promise for crowd navigation, due to its ability to model complex decision-making processes \cite{Chen2017, Chen2018, Everett2018, Chen2019, Chen2020}. In learning-based methods, the crowd navigation problem is typically formulated as a Markov decision process, where a DRL agent uses a neural network to select the next action, either by directly predicting the optimal action or by choosing one that maximizes the expected value of state at the next time step.

\begin{figure}[t]
    \centering
    \includegraphics[width = \linewidth]{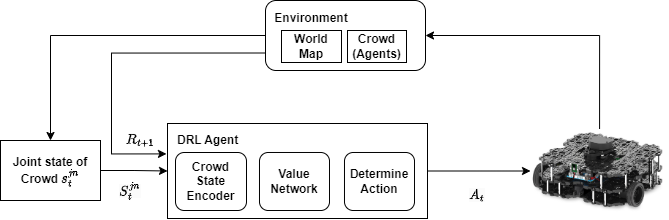}
    \caption{Framework for collision avoidance using deep reinforcement learning}
    \label{fig:DRL SRN}
\end{figure}

The neural network employed by the deep reinforcement learning (DRL) agent for social navigation typically consists of two main components: the crowd state encoder and the value/policy network (see \autoref{fig:DRL SRN}). The crowd state encoder is responsible for the spatial processing of crowd data, mapping all observable agent information to a latent space. This approach is designed to accommodate a variable number of agents \cite{Everett2018} and to capture the interactions between agents and the robot \cite{Chen2019}. Most research in social robot navigation (SRN) using DRL has concentrated on improving crowd state encoder strategies, and comparatively little attention has been directed toward the value network. In current implementations, the value network is modeled as a simple multi-layer perceptron.

In recent years, deep learning-based state space models (SSM) \cite{S4-gu2021,gu2023} have shown to be powerful tools for sequence modelling tasks such as audio generation, natural language processing, and DNA sequencing. ``Mamba"  is a new deep learning-based State Space Model (SSM) that has achieved results comparable to transformers in sequence modelling tasks \cite{gu2023}.

This paper presents a novel approach for collision avoidance using deep reinforcement learning, CAMRL (Context Aware Mamba-based Reinforcement Learning). Our main contribution here is introducing temporal processing of crowd states (joint state of the robot and all the other agents) by utilizing Mamba's ability to model sequences. We also identified gaps in the simulation on which previous models were evaluated on, and hence used a more rigorous testing approach \cite{Sigal2023} to evaluate the effectiveness of our model. 

\section{The collision avoidance paradigm}\label{sec:collision avoidance}
The state of the agents in the context of non-communicating multi-agent collision can be defined as the concatenation of its observable state and its hidden state. For each agent, the position \(\mathbf{p = [p_x, p_y]}\), velocity \(\mathbf{v = [v_x, v_y]}\), and the radius of the bounding circle of the agent, \(r\), can be observed. The intended goal position \(\mathbf{p_g = [p_{gx},p_{gy}]}\), and the preferred velocity \(\mathbf{v_{pref}}\) are hidden. Thus, the problem can be formulated as a partially observable sequential decision-making problem. In this paper, the hidden state of an agent would be sometimes called the intention of the robot. Furthermore, the joint state of the system denotes 
the concatenation of the complete state of the robot and the observable state of the other agents, i.e. for a system with n+1 agents the joint state is given by \(s_t^{jn} = [s_t, \tilde{s}_t^{o,1}, \tilde{s_t}^{o,2}, ..... , \tilde{s}_t^{o,n}]\), where the observable state of \(i_{th}\) other agent is given by \(\tilde{s}_t^{o,i}\).

The goal of a collision avoidance algorithm is to reduce the expected time to reach the goal while avoiding any collisions. The n-agent collision avoidance problem is mathematically formulated following the approach presented in \cite{Chen2017}. 

\begin{subequations}
\label{eq: Problem Formulation}
\begin{align}
    argmin_{\pi(s_t^{jn})}   & \mathbb{E}[t_g|s_0^{jn}, \pi,\tilde{\pi}_1,...\tilde{\pi}_n] \label{expectation}\\
    s.t.  \                       & \forall \ t,i:\| p_{t} - \tilde{p}_{i,t}\|_2 \geq r+ \tilde{r_i} \label{collision}\\
                                & p_{end} = p_g \label{goal contraint}\\
                                & p_t = {p}_{t-1}+\Delta t \ \pi(s_{0:t}^{jn}) \nonumber \\
                                & \tilde{p}_{i,t} = \tilde{p}_{i,t-1}+\Delta t \ \tilde{\pi}_i(\tilde{s}_{0:t}^{jn}) \label{agents kinematics},
\end{align}
\end{subequations}
where \(t_g\) is the time to reach the goal, \(s_t^{jn}\) is the joint state of the system at time \textit{t}, \(\tilde{s}_t^{jn}\) is the joint state at time \textit{t} of the system concerning another agent, \(\pi\) and \(\tilde{\pi}_i\) are the robot and \(i^{th}\) agents policy (a function that maps the states to actions), \(p_g\) is the goal position, \(p_t\), \(\tilde{p_{t,i}}\) is the position of the robot and the \(i^{th}\) agent at time \(t\), \(r\) and \(r_i\) is the radius of the robot and the \(i^{th}\) agent, \(\Delta t\) is the time step and \(p_{end}\) is the position of the robot at the end of the current episode. \autoref{expectation} is the expectation constraint, i.e. the robot's policy should minimize the expected time to reach the goal. \autoref{collision} is the collision constraint that the robot should not collide with the \(i^{th}\) agent. \autoref{goal contraint} is the goal constraint, i.e. the robot is at the goal position at the end of the episode

\section{Related Work}
This section reviews some basic and relevant ideas important to deep reinforcement learning-based social robot navigation solutions and the background for Mamba and state space models.

\subsection{DRL models for collision avoidance}
A number of DRL-based collision avoidance algorithms have been proposed to date. Some of these algorithms along with their brief descriptions are included here for reference.
   \begin{enumerate}
       \item \textbf{CADRL}: It uses a simple multi-layer perception (MLP) to approximate the value function \cite{Chen2017}. In this model the crowd encoding layer does not exist and the value network is operating on joint state.
       \item \textbf{GA3C CADRL (LSTM-RL)}: It uses a long short-term memory (LSTM) to compress all the observable information to a fixed-length vector before passing it into an MLP block \cite{Everett2018}.
       \item  \textbf{SARL}: It uses a social attention network on top of an MLP, assigning a weight to each pedestrian in the crowd \cite{Chen2019}.
       \item \textbf{LM-SARL}: It uses a local map as an input to the social attention network. This is done to model human-to-human interactions as well \cite{Chen2019}.
       \item \textbf{Relational Graph learning}: It uses graph convolution neural networks to learn interaction between all the agents in the environment \cite{Chen2020}.
   \end{enumerate}

As noted in \cite{Sigal2023}, algorithms trained and tested on simple circle crossing environments (see \autoref{fig:div-4}), which has been the case in \cite{Chen2017, Chen2018, Everett2018, Chen2019, Chen2020}, often fail to generalize to more complex, real-world scenarios. This limitation arises because such environments do not accurately simulate the dynamics of real-world crowds \cite{thrun2002probabilistic}. To address this, \cite{Sigal2023} has proposed improved training schemes using more diverse datasets, which show promise and will also be utilized in this work. Nevertheless, further investigation is needed to assess how effectively these simulated datasets replicate real-world crowd behaviors.

\subsection{State Space Models}
State space models (SSM) have been widely used to describe dynamical systems in econometrics and finance \cite{hamilton1994state},
control systems \cite{aastrom2021feedback}, robotics \cite{thrun2002probabilistic}, etc., using first-order differential equations. The standard state-space model in continuous time is described by \autoref{eq:continuous state space} and in discrete-time by the recurrence relations in \autoref{eq:discrete SSM}, where $x$ is the state, $u$ is the input signal/sequence and $y$ is the output signal/sequence. 

\begin{align}
    x'(t) &= Ax(t) + Bu(t) \nonumber\\
    y(t)  &= Cx(t) + Du(t)
    \label{eq:continuous state space}
\end{align}

where \(A,B,C,\text{ and } D\) are matrices of appropriate sizes.

The continuous form of the state space model can be discretised using a fixed step size \(\Delta\) and any discretization technique such as zero order hold sampling.
\begin{align}
    x_n &= \Bar{A}x_{n-1} + \Bar{B}u_n \nonumber\\
    y_n &= \Bar{C}x_n + \Bar{D}u_n
    \label{eq:discrete SSM}
\end{align}
The discrete form of SSM can also be represented using a convolutional view\cite{gu2021_LSSL} :
\begin{align}
    y &= \Bar{K} \mathop{*}u \\
    \Bar{K} &\in \mathbb{R} := K_L(\Bar{A},\Bar{B},\Bar{C}):=(\Bar{CA^iB})_{i\in [L]}
    \label{eq:conv SSM}
\end{align}

where \([L]\) is an indexing set defined over the input sequence and \(\Bar{K}\) is called the SSM convolutional filter. The convolutional representation of SSM is relevant because now output sequence \(y\) given the input sequence \(u\) can be computed efficiently using Fast Fourier Transform \cite{S4-gu2021}.

\subsection{State Space Models in Learning}
Basic state-space models (SSMs) often perform poorly when used as a model in deep learning applications, where matrices $A$, $B$, $C$, and $D$ are learned. However, recent work \cite{gu2021_LSSL, gu2023, gu2021efficiently, gu2022HIPPO_TRAIN} has achieved significant success with SSMs by imposing special structures on the state matrix $A$. One such example is the structured state-space model (S4) \cite{gu2021efficiently}, where the matrix $A$ is designed as a normal plus low-rank matrix, initialized using the HiPPO matrix. The HiPPO matrix provides an optimal solution to an online function approximation problem \cite{gu2020hippo}, and models utilizing HiPPO have shown remarkable performance on tasks involving long-range dependencies, outperforming transformers on the PathBench test suite. Additionally, due to the convolutional representation of SSMs, they can be parallelized, offering faster training times compared to recurrent neural networks \cite{gu2021efficiently}. And with their recurrent representation of SSMs they can infer faster than a transformer (which scales quadratically with the input sequence).

\subsubsection{Selective State Space Model (Mamba)}
An addition to the S4 model, called Mamba, was presented by Gu et al. in \cite{gu2023}. Unlike S4, which uses time-invariant parameters, Mamba proposes selective SSMs through input-dependent parameterization, enabling the model to selectively propagate or forget information based on the current token. This allows the model to use only the relevant information from the input sequence. The algorithm for S6 (Mamba) is given in \textbf{Algorithm \ref{alg:ssm_selection}}.

\begin{algorithm}[H]
\caption{Mamba (S6) \cite{gu2023}}
\label{alg:ssm_selection}
\textbf{Input:} $x: (B, L, D)$ \newline
\textbf{Output:} $y: (B, L, D)$ \newline
$\textbf{A:} (D, N) \leftarrow Parameter$ \{Represents structured \(N \times N\) matrix\} \newline
$\textbf{B:} (B, L, N) \leftarrow s_B(X)$  
\newline
$\textbf{C:} (B, L, N) \leftarrow s_C(x)$ 
\newline
\textbf{$\Delta:$} $(B,L,D) \leftarrow \tau_{\Delta}(Parameter + s_{\Delta}(x))$ 
\newline
$\overline{A}, \overline{B}: (B, L, D, N) \leftarrow discretize (\Delta, A, B)$ 
\newline
$y \leftarrow SSM(\overline{A}, \overline{B}, C)(x)$ 
\{Time varying: recurrence(scan) only\} \newline
\textbf{return} $y$
\end{algorithm}

\subsubsection{State Space Models in the Context of Reinforcement Learning}
A reinforcement learning algorithm that is capable of long-term planning is one of the central goals of reinforcement learning (RL) \cite{david2022decision}. The ability of DSSM models to handle long-range dependencies and handle large contexts opens new possibilities in RL. Due to the cutting-edge nature of DSSM, little work has been done to investigate its efficacy in reinforcement learning. However, the work that has been done indicates that DSSM is a promising avenue for RL. For example, \cite{lu2024structureds5} used S5 (a variant of S4) for online meta-reinforcement learning where S5 outperformed LSTMs \cite{duan2016metaRL}, while \cite{david2022decision} showed an offline implementation of S4 and showed that S4 outperformed transformers in various RL tasks. In this paper, we are adopting this approach for reinforcement learning and testing it for social robot navigation.

\section{Methodology}\label{sec:methodology}
In this section, we first discuss the DRL algorithm used, the reward function, and how it connects to social robot navigation, SSMs in the context of DRL, and then explain our model CAMRL (Context-Aware Mamba-based Reinforcement Learning) in detail.
\subsection{Preliminaries}
\subsubsection{Deep V-learning}
Deep V-learning, a deep reinforcement learning framework based on the Deep Q-Network algorithm but designed for continuous action spaces, was first introduced in \cite{Chen2017} for the development of CADRL. It has two learning steps:
\begin{itemize}
    \item \textbf{Imitation Learning:} The value network is trained on the training dataset.
    \item \textbf{Reinforcement Learning:}  Iteratively improves the value network through experience replay and target network updates.
\end{itemize}

\subsubsection{Reward Function}
The reward function is modelled as follows:
\begin{equation}
R_1(s, a_t) =
\begin{cases}
-0.25, & \text{if } \mathbf{d_t} \leq 0 \\
 \dfrac{(\mathbf{d_t} -\mathbf{r_c})\Delta t}{2}, & \text{else if } \mathbf{d_t} < \mathbf{r_c}  \\
1, & \text{else if } \mathbf{p_t} = \mathbf{p_g} \\
-0.5,&\text{else if }\mathbf{t} = \mathbf{25} \\
0, & \text{otherwise}

\end{cases}
\end{equation}
where \(\mathbf{d_t}\) is the separation distance between
robot and the humans, \(\mathbf{p_t}\) is robot's current position and \(\mathbf{p_g}\) is the goal position, and \(\mathbf{r_c}\) is the radius of discomfort. \\
This reward function encourages the robot to avoid collisions, not violate the radius of discomfort of a human, and to finish the navigation task in a set time. 

\subsubsection{State space models in the context of reinforcement learning}
The discrete representation can be thought
of as sampling from a continuous signal. The authors reported that deep-state space models are good at handling long-range dependencies. Given that the optimal value function is a causal system, i.e. dependent on previous states: 
\begin{equation}
    V^8(\mathbf{s^{jn}_0}) = \sum^{T}_{k =0}[{ \gamma^{t}R(\mathbf{s^{jn}_t}, \pi^*(\mathbf{s^{jn}_t})}],
\end{equation}
 where \(V(s^{jn}_t)\) is the value of the joint state at time \(t\), \(R_t\) is the reward at time \(t\), \(\gamma \in [0,1)\) is a discount factor, and  $\pi^*(\mathbf{s^{jn}_t})$ is the optimal policy \cite{Chen2017}.

Hence, it makes sense to use a neural network that can model a sequence of states instead of a simple multi-layered perceptron.

\subsection{Context Aware Mamba-based Reinforcement Learning}
\begin{figure}[t]
    \centering
        \includegraphics[width = \linewidth]{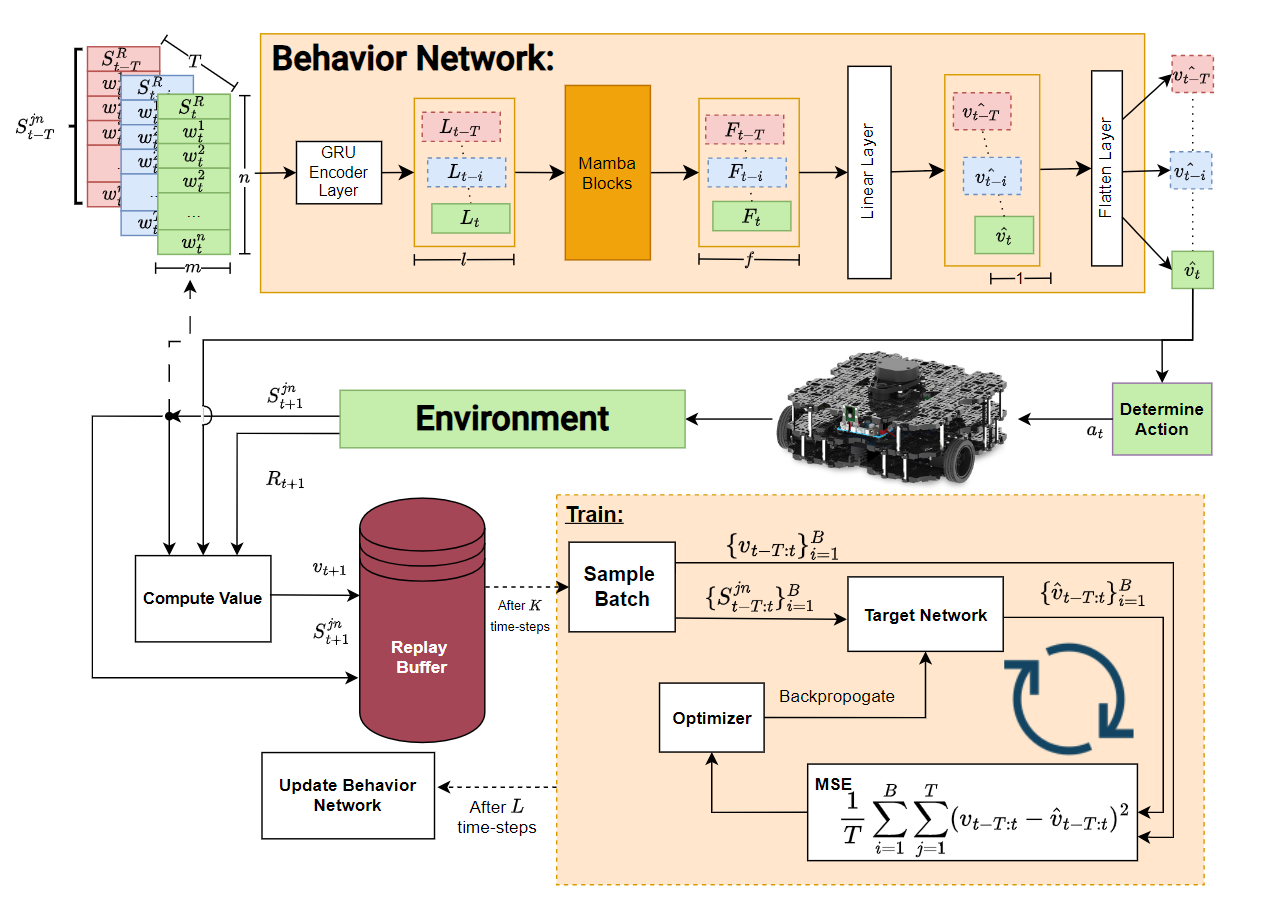}
    \caption{Architecture for CAMRL}
    \label{fig:architecture}
\end{figure}

Leveraging the ability of SSMs to handle long-range dependencies effectively, our proposed model, Context-aware Mamba-based Reinforcement Learning (CAMRL), is provided with a sequence of states at the present and past times. This is different from most previous works that provide the current state to the value network.
The CAMRL model, as illustrated in \autoref{fig:architecture}, can be divided into two primary components: the crowd state encoder and the value network. The crowd state encoder uses a GRU encoder to process a temporal slice of joint states, that will be referred to as the temporal crowd state vector, i.e. the joint state of the crowd from time \( t-T \) to \( t \), i.e. \( {s_{t-T:t}^{jn}}\). This encoder functions as a mechanism that compresses all the information of the crowd at a particular time instant into a fixed-length vector \( L_t \) within a latent space, referred to as the crowd space which has all configurations of a crowd. 

Following the encoding process, the output of the crowd state encoder is fed into the value network. This value network is modeled using a Mamba block, which consists of four stacked Mamba layers. These vectors are then processed through the Mamba block to produce the temporal value vector \( v_{k =t-T}^t \), i.e. a vector containing the values at time \( t-T \) to \( t \). The final value at time \( t \) is used to determine the action \( a_t \) for the environment. The action that maximizes value at time \(t\) is selected.

During training, a replay buffer stores the temporal crowd state and value vectors, which are used to train the target network through backpropagation and optimization based on the mean squared error (MSE) loss on a batch of size \textit{B} of the temporal value vector (\(\{v_{t-T:t}\}^B_{i=1}\)) and the predicted batch of temporal value vectors (\(\{\hat{v}_{t-T:t}\}^B_{i=1}\)). After every \textit{K} episode the behavior network is updated by the target network. 

This systematic approach allows CAMRL to effectively learn and predict actions based on long-range dependencies, ensuring robust performance in dynamic and complex environments.

\section{Experimental Setup}\label{sec:exp-setup}
The developed CAMRL model was tested for social-robot navigation in simulation. The training and testing setup is outlined in this section. Metrics used for comparing our model with existing models are described in \autoref{metrics}.

\subsection{Environment setup}\label{envconf}
The typical setup for training and testing employed in literature \cite{Chen2017, Chen2018, Everett2018, Chen2019, Sigal2023} is shown in Figure 3(a). The basic idea is as follows: humans (other agents), indicated by numbered circles, start at a random position on the perimeter of a circle or square and their 
goal position is set roughly on the opposite side of the shape. The
other agents try to navigate to their end goals using ORCA \cite{orca}. The robot, represented by the yellow circle, follows the algorithm-determined policy to navigate to its goal, the red star, avoiding any collisions. The robot is only aware of the variables
in the observable state detailed in \autoref{sec:collision avoidance}. The robot is not visible to the humans during testing or training, ensuring the trained policy does not have the robot forcing the other agents to change their trajectories to allow the robot to achieve a high reward \cite{Liu2023}.

However, the simplistic circle crossing setting does not generalize well to real-world scenarios \cite{Survey2023-core-challenges}, \cite{Sigal2023}. Thus \cite{Sigal2023} proposed the Diverse-4 dataset, introducing two additional environments, dense and large, that take into account the crowd density and the distance the robot has to cover. It also recommends SFM as the model for human agents for more rigorous testing of the policy.

The test environments are divided into simple (or baseline), dense, and large. The simple environment is small with a crowd density of 0.1 humans per m$^2$, and is typically used in the literature. The dense environment has the same area but doubles the crowd density, simulating navigation in challenging conditions. The large environment covers a larger area while maintaining the baseline density; this allows for evaluating the model's long-horizon task capabilities.

\subsection{Training and Testing Setup}\label{training}
Our model was trained in the baseline circle crossing environment used by previous models like CADRL \cite{Chen2017}, LSTM-RL \cite{Everett2018}, and SARL \cite{Chen2019} with pedestrians modelled by ORCA only. 

\begin{figure}
\centering
\begin{tabular}{cc}
  \subcaptionbox{Baseline Circle Crossing (4m radius, 5 agents)}[0.4\linewidth]{\includegraphics[width=\linewidth]{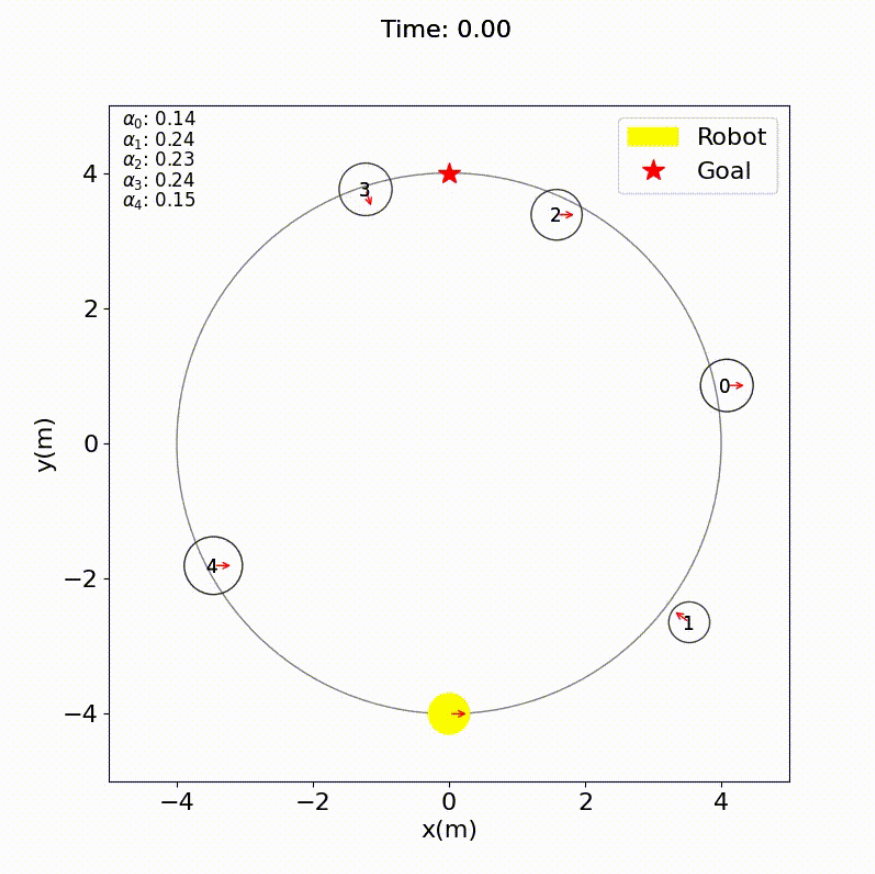}
    } 
  &
  \subcaptionbox{Baseline Square Crossing (10m wide, 10 agents)}[0.4\linewidth]{\includegraphics[width=\linewidth]{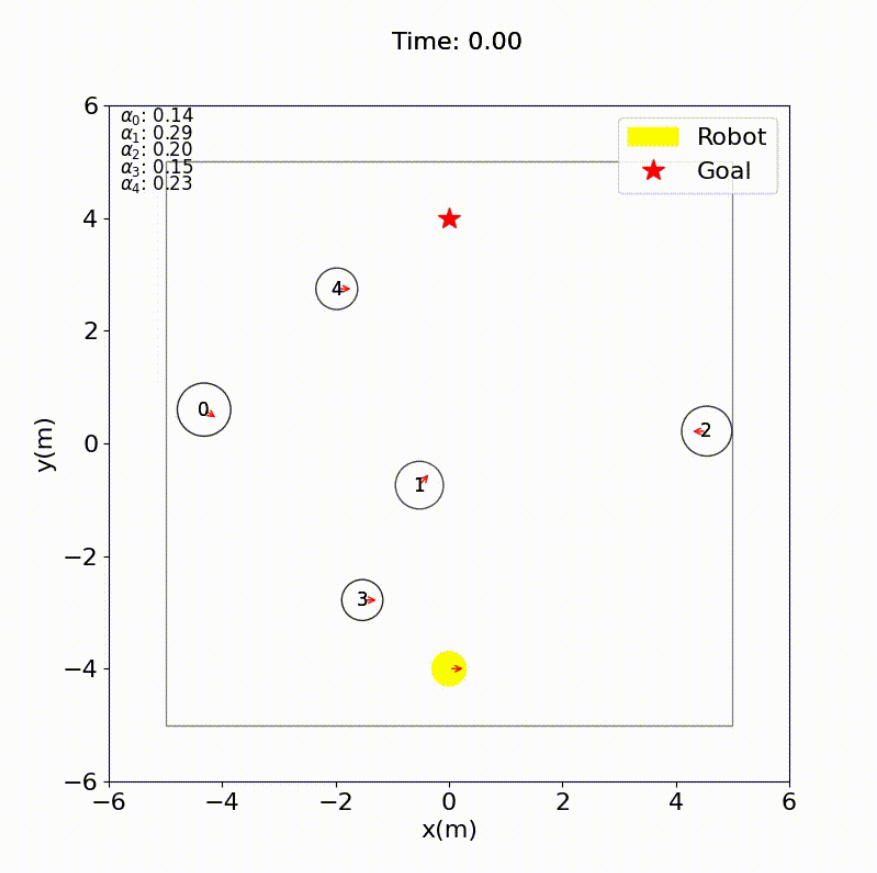}} \\

  \subcaptionbox{Dense Circle Crossing (4m radius, 10 agents)}[0.4\linewidth]{\includegraphics[width=\linewidth]{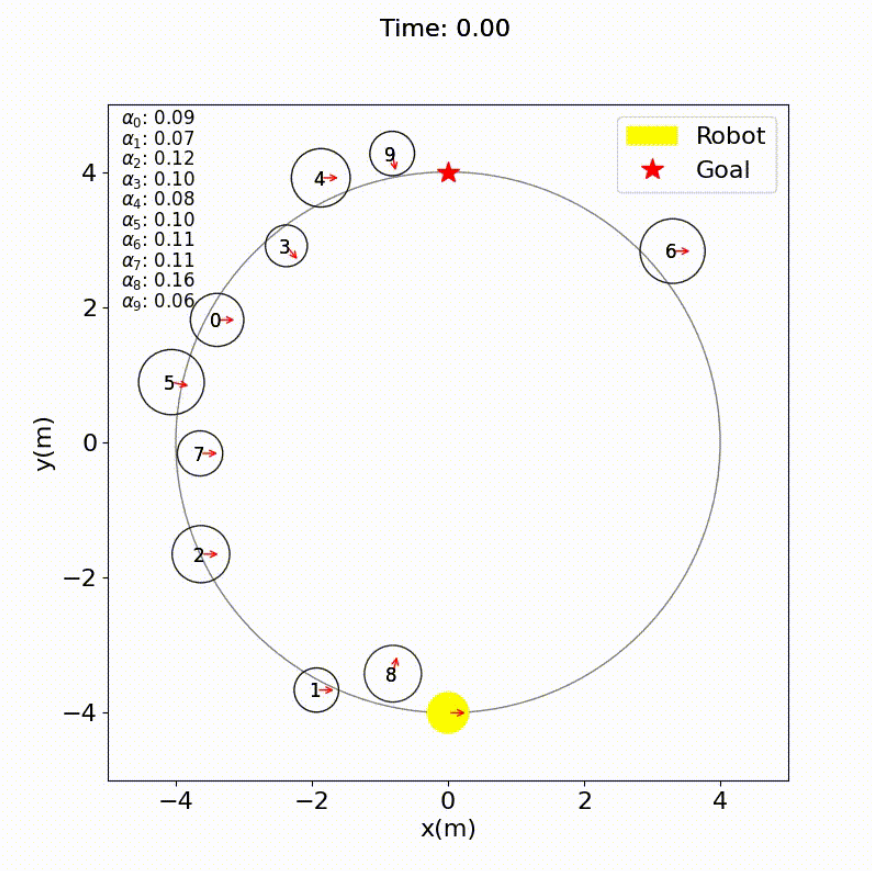}} &
  \subcaptionbox{Dense Square Crossing (10m wide, 20 agents)}[0.4\linewidth]{ \includegraphics[width=\linewidth]{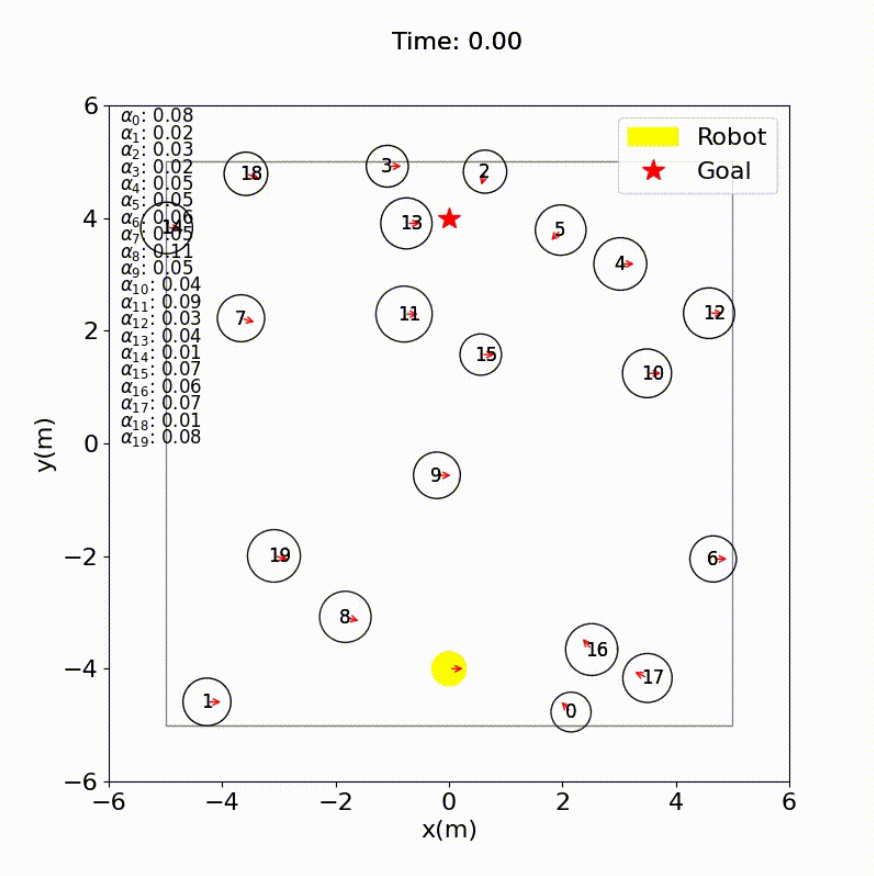}} \\

  \subcaptionbox{Large Circle Crossing (6m radius, 12 agents)}[0.4\linewidth]{\includegraphics[width=\linewidth]{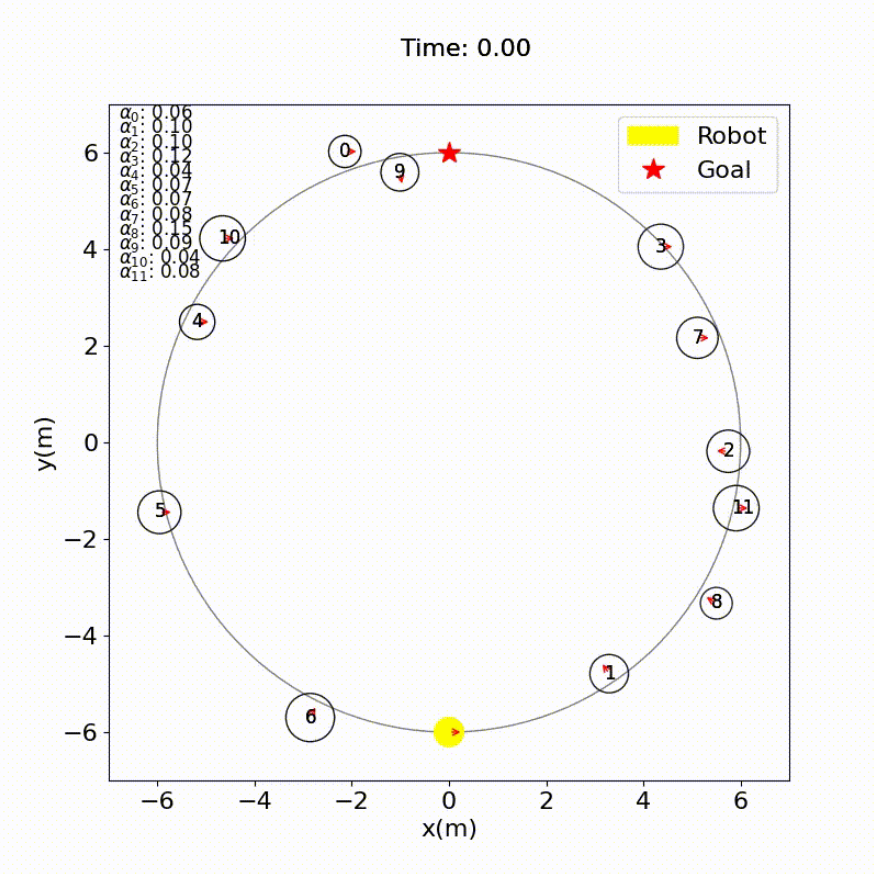}} &
  \subcaptionbox{Large Square Crossing (14m wide, 20 agents)}[0.4\linewidth]{\includegraphics[width=\linewidth]{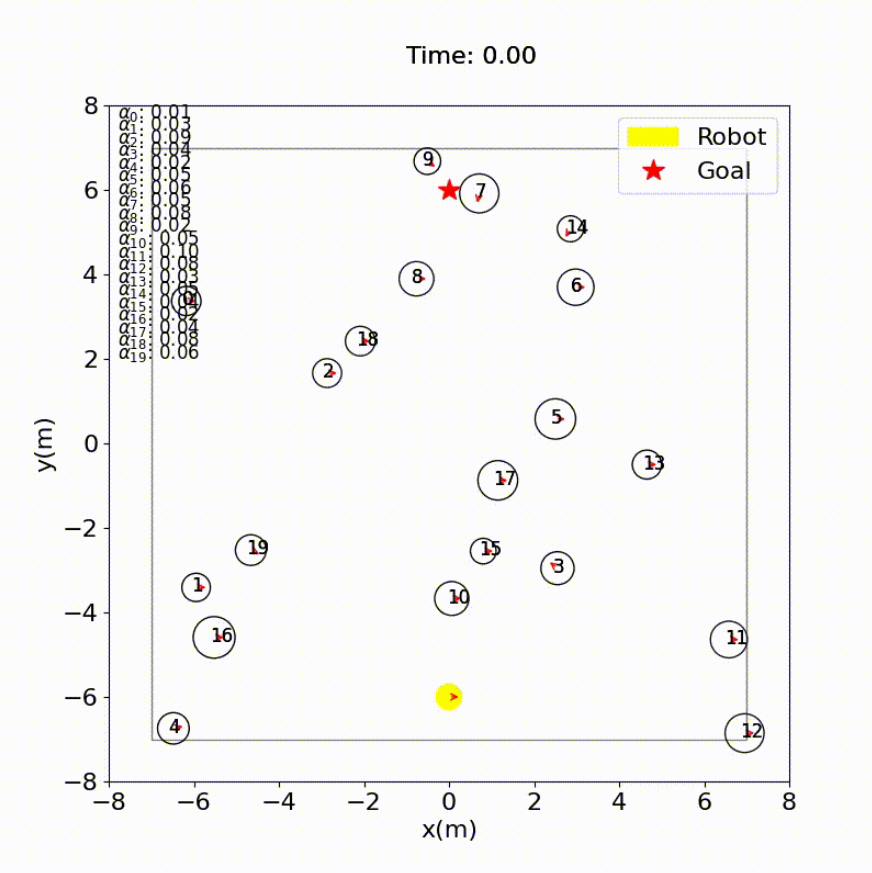}} 
\end{tabular}
\caption{Diverse-4 Testing Scenarios from \cite{Sigal2023}.}\label{fig:div-4}
\end{figure}

Based on the recommendations in \cite{Sigal2023}, our model was tested in six testing environments, namely, baseline circle and square crossing, dense circle and square crossing, large circle and square crossing (as shown in \autoref{fig:div-4}) with 50 test cases in each environment. These environments were chosen to span diverse crossing types, crowd densities, and task lengths. 

Additionally, we used both ORCA and SFM for crowd modelling in tests, which allowed for a more realistic simulation of humans \cite{Survey2023-core-challenges}.

For training, the models learn their policy over a number of episodes. In each episode, the robot tries to reach the goal from its start position while trying to avoid other agents. An episode is terminated when the robot either reaches the goal, collides with an agent or does not make it to the goal within the given time (times out). In our case, the time out is 25 seconds. 

We trained and tested our model using a 32 GB Titan RTX with an Intel Core i7-9700K CPU at 3.60GHz. Mamba utilizes CUDA functionalities and hence requires a GPU; however, the other models can be trained using a CPU only.

\subsection{Metrics}\label{metrics}
In order to evaluate the performance of our model we used standard metrics commonly used for comparing DRL models in the literature. 
We compared our model with three state-of-the-art models found in the literature: CADRL \cite{Chen2017}, LSTM-RL \cite{Everett2018}, and SARL \cite{Chen2019}. The metrics are defined in \autoref{table:evaluation criteria}.

\begin{table}[t]
\centering
\caption{Metrics identified for Evaluation}
\label{table:evaluation criteria}
\begin{tabular}{|p{2.5cm}|p{5.9cm}|}
\hline
\textbf{Metrics} & \textbf{Definition}
\\
\hline
Success Rate & The amount of times the robot reaches the goal without collisions and within the given time frame in the given number of episodes. \\
\hline
Collision Rate & The amount of times the robot collides with an agent or an obstacle in the given number of episodes. \\
\hline
Timeout Rate & The amount of times the robot does not reach the goal within the given time and avoids collisions in the given number of episodes. \\
\hline
Time taken (s) & The average time the robot takes to reach the goal. \\
\hline
Discomfort Frequency & The number of times the robot got too close to other agents in the given number of episodes. \\
\hline
Discomfort Distance (m) & The average distance the robot maintained with other agents. \\ 
\hline
\end{tabular}
\end{table}

\section{Results}\label{results}
\subsection{Comparison of CAMRL with other models}\label{comparison}
The results of testing are summarized in \autoref{table:comparison-table}. 

\begin{table*}[t]
\centering
\caption{Comparison of our model with state-of-the-art models on all metrics}\label{table:comparison-table}
\begin{tabular}{|l|*{6}{c|}}  
\hline
\textbf{Policy} & \textbf{Success} & \textbf{Collision} & \textbf{Timeout} & \textbf{Time taken (s)} & \textbf{Disc. Freq} & \textbf{Disc. Dist. (m)} \\ \hline
ORCA & 0.51 & 0.49 & \textbf{0.00} & $12.26\pm2$ & 0.24 & $0.09\pm0.1$ \\
CADRL & 0.73 & 0.16 & 0.11 & $16.04\pm4$ & 0.10 & $0.13\pm0.06$ \\
LSTM-RL & 0.56 & 0.38 & 0.06 & $15.75\pm5$ & 0.21 & $0.09\pm0.06$ \\
SARL & 0.83 & 0.17 & \textbf{0.00} & \textbf{11.20$\pm$1} & 0.21 & $0.13\pm0.05$ \\
CAMRL & \textbf{0.88} & \textbf{0.11} & 0.01 & $13.91\pm2$ & \textbf{0.07} & \textbf{0.16$\pm$0.04} \\
\hline
\end{tabular}
\end{table*}

\begin{figure}[t]
    \centering
    \includegraphics[width=\linewidth]{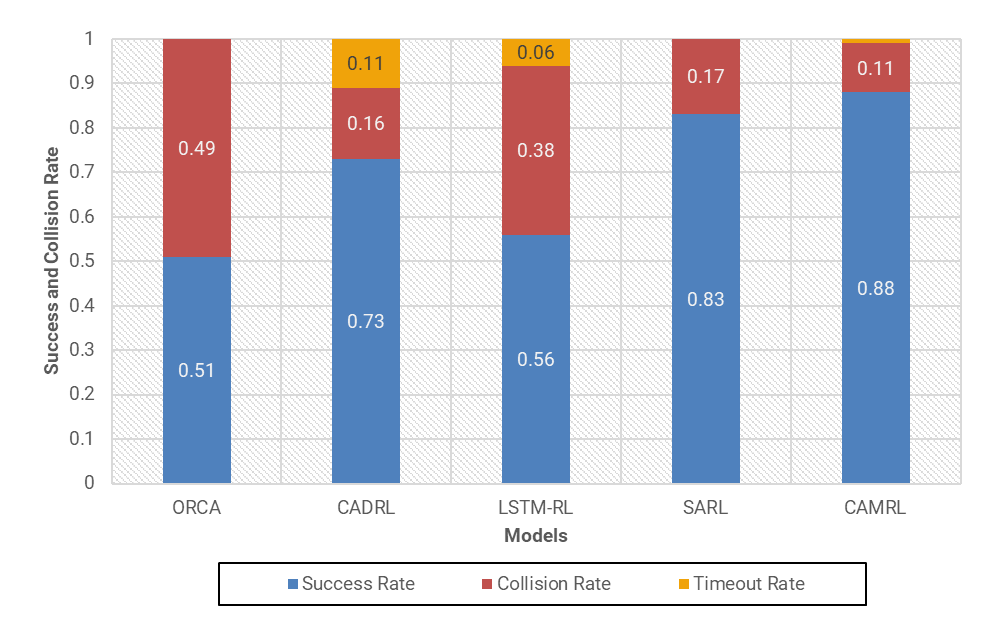}
    \caption{Comparison of our model with state-of-the-art models on the basis of success, timeout and collision rate}
    \label{comparison-graph}
\end{figure}
Compared to CADRL, LSTM-RL and SARL, our model, CAMRL, performs the best. It gives a 5\% higher success rate than SARL, which is the best out of the three. CAMRL also has the least number of collisions and manages to bring down the collision rate by a margin of 6\% compared to SARL. CAMRL has a timeout rate of 1\%, where it does not reach the goal in the given time but manages to avoid collisions. This weakness is covered up by the higher success rate and lower collision rate compared to the other three models, for scenarios where avoiding collisions might be a more preferred outcome than the robot not reaching its end destination in time. These results can also be seen in \autoref{comparison-graph}.
Additionally, CAMRL maintains a greater distance from other agents, on average, compared to other models. Overall, it performed better in maintaining comfort distance by maintaining discomfort frequency at 7\%, the least compared to its counterparts. 

\section{Conclusion}
This work proposes CAMRL, a new model for path planning in social robot navigation (SRN). CAMRL performed either comparably or better than current state-of-the-art solutions across a rigorous set of test cases. This work also demonstrates that state space modelling offers promising solutions for path planning in SRN. Following this, future work could involve comparing CAMRL with graph learning solutions \cite{Chen2020}\cite{Liu2023} and implementing CAMRL on a real-world robot.  

\bibliographystyle{IEEEtran}
\bibliography{main}

\begin{thebibliography}{10}
\providecommand{\url}[1]{#1}
\csname url@samestyle\endcsname
\providecommand{\newblock}{\relax}
\providecommand{\bibinfo}[2]{#2}
\providecommand{\BIBentrySTDinterwordspacing}{\spaceskip=0pt\relax}
\providecommand{\BIBentryALTinterwordstretchfactor}{4}
\providecommand{\BIBentryALTinterwordspacing}{\spaceskip=\fontdimen2\font plus
\BIBentryALTinterwordstretchfactor\fontdimen3\font minus \fontdimen4\font\relax}
\providecommand{\BIBforeignlanguage}[2]{{%
\expandafter\ifx\csname l@#1\endcsname\relax
\typeout{** WARNING: IEEEtran.bst: No hyphenation pattern has been}%
\typeout{** loaded for the language `#1'. Using the pattern for}%
\typeout{** the default language instead.}%
\else
\language=\csname l@#1\endcsname
\fi
#2}}
\providecommand{\BIBdecl}{\relax}
\BIBdecl

\bibitem{orca}
J.~Van Den~Berg, S.~J. Guy, M.~Lin, and D.~Manocha, ``Reciprocal n-body collision avoidance,'' in \emph{Robotics Research: The 14th International Symposium ISRR}.\hskip 1em plus 0.5em minus 0.4em\relax Springer, 2011, pp. 3--19.

\bibitem{sfm}
D.~Helbing and P.~Molnar, ``Social force model for pedestrian dynamics,'' \emph{Physical review E}, vol.~51, no.~5, p. 4282, 1995.

\bibitem{Chen2017}
Y.~F. Chen, M.~Liu, M.~Everett, and J.~P. How, ``Decentralized non-communicating multiagent collision avoidance with deep reinforcement learning,'' in \emph{2017 IEEE international conference on robotics and automation (ICRA)}.\hskip 1em plus 0.5em minus 0.4em\relax IEEE, 2017, pp. 285--292.

\bibitem{Chen2018}
Y.~F. Chen, M.~Everett, M.~Liu, and J.~P. How, ``Socially aware motion planning with deep reinforcement learning,'' in \emph{2017 IEEE/RSJ International Conference on Intelligent Robots and Systems (IROS)}.\hskip 1em plus 0.5em minus 0.4em\relax IEEE, 2017, pp. 1343--1350.

\bibitem{Everett2018}
M.~Everett, Y.~F. Chen, and J.~P. How, ``Motion planning among dynamic, decision-making agents with deep reinforcement learning,'' in \emph{2018 IEEE/RSJ International Conference on Intelligent Robots and Systems (IROS)}.\hskip 1em plus 0.5em minus 0.4em\relax IEEE, 2018, pp. 3052--3059.

\bibitem{Chen2019}
C.~Chen, Y.~Liu, S.~Kreiss, and A.~Alahi, ``Crowd-robot interaction: Crowd-aware robot navigation with attention-based deep reinforcement learning,'' in \emph{2019 international conference on robotics and automation (ICRA)}.\hskip 1em plus 0.5em minus 0.4em\relax IEEE, 2019, pp. 6015--6022.

\bibitem{Chen2020}
C.~Chen, S.~Hu, P.~Nikdel, G.~Mori, and M.~Savva, ``Relational graph learning for crowd navigation,'' in \emph{2020 IEEE/RSJ International Conference on Intelligent Robots and Systems (IROS)}.\hskip 1em plus 0.5em minus 0.4em\relax IEEE, 2020, pp. 10\,007--10\,013.

\bibitem{S4-gu2021}
\BIBentryALTinterwordspacing
A.~Gu, K.~Goel, and C.~R{\'{e}}, ``Efficiently modeling long sequences with structured state spaces,'' \emph{CoRR}, vol. abs/2111.00396, 2021. [Online]. Available: \url{https://arxiv.org/abs/2111.00396}
\BIBentrySTDinterwordspacing

\bibitem{gu2023}
A.~Gu and T.~Dao, ``Mamba: Linear-time sequence modeling with selective state spaces,'' \emph{arXiv preprint arXiv:2312.00752}, 2023.

\bibitem{Sigal2023}
A.~Sigal, H.-C. Lin, and A.~Moon, ``Improving reinforcement learning training regimes for social robot navigation,'' \emph{arXiv preprint arXiv:2308.14947}, 2023.

\bibitem{thrun2002probabilistic}
S.~Thrun, ``Probabilistic robotics,'' \emph{Communications of the ACM}, vol.~45, no.~3, pp. 52--57, 2002.

\bibitem{hamilton1994state}
J.~D. Hamilton, ``State-space models,'' \emph{Handbook of econometrics}, vol.~4, pp. 3039--3080, 1994.

\bibitem{aastrom2021feedback}
K.~J. {\AA}str{\"o}m and R.~Murray, \emph{Feedback systems: an introduction for scientists and engineers}.\hskip 1em plus 0.5em minus 0.4em\relax Princeton university press, 2021.

\bibitem{gu2021_LSSL}
A.~Gu, I.~Johnson, K.~Goel, K.~Saab, T.~Dao, A.~Rudra, and C.~R{\'e}, ``Combining recurrent, convolutional, and continuous-time models with linear state space layers,'' \emph{Advances in neural information processing systems}, vol.~34, pp. 572--585, 2021.

\bibitem{gu2021efficiently}
A.~Gu, K.~Goel, and C.~Re, ``Efficiently modeling long sequences with structured state spaces,'' in \emph{International Conference on Learning Representations}, 2021.

\bibitem{gu2022HIPPO_TRAIN}
A.~Gu, I.~Johnson, A.~Timalsina, A.~Rudra, and C.~R{\'e}, ``How to train your hippo: State space models with generalized orthogonal basis projections,'' \emph{arXiv preprint arXiv:2206.12037}, 2022.

\bibitem{gu2020hippo}
A.~Gu, T.~Dao, S.~Ermon, A.~Rudra, and C.~R{\'e}, ``Hippo: Recurrent memory with optimal polynomial projections,'' \emph{Advances in neural information processing systems}, vol.~33, pp. 1474--1487, 2020.

\bibitem{david2022decision}
S.~B. David, I.~Zimerman, E.~Nachmani, and L.~Wolf, ``Decision s4: Efficient sequence-based rl via state spaces layers,'' in \emph{The Eleventh International Conference on Learning Representations}, 2022.

\bibitem{lu2024structureds5}
C.~Lu, Y.~Schroecker, A.~Gu, E.~Parisotto, J.~Foerster, S.~Singh, and F.~Behbahani, ``Structured state space models for in-context reinforcement learning,'' \emph{Advances in Neural Information Processing Systems}, vol.~36, 2024.

\bibitem{duan2016metaRL}
Y.~Duan, J.~Schulman, X.~Chen, P.~L. Bartlett, I.~Sutskever, and P.~Abbeel, ``Rl2: Fast reinforcement learning via slow reinforcement learning,'' \emph{arXiv preprint arXiv:1611.02779}, 2016.

\bibitem{Liu2023}
S.~Liu, P.~Chang, Z.~Huang, N.~Chakraborty, K.~Hong, W.~Liang, D.~L. McPherson, J.~Geng, and K.~Driggs-Campbell, ``Intention aware robot crowd navigation with attention-based interaction graph,'' in \emph{2023 IEEE International Conference on Robotics and Automation (ICRA)}.\hskip 1em plus 0.5em minus 0.4em\relax IEEE, 2023, pp. 12\,015--12\,021.

\bibitem{Survey2023-core-challenges}
C.~Mavrogiannis, F.~Baldini, A.~Wang, D.~Zhao, P.~Trautman, A.~Steinfeld, and J.~Oh, ``Core challenges of social robot navigation: A survey,'' \emph{ACM Transactions on Human-Robot Interaction}, vol.~12, no.~3, pp. 1--39, 2023.

\end{thebibliography}
\end{document}